\documentclass{article}

\PassOptionsToPackage{numbers, compress}{natbib}

\usepackage[preprint]{neurips_2021}

\usepackage[utf8]{inputenc} % allow utf-8 input
\usepackage[T1]{fontenc}    % use 8-bit T1 fonts
\usepackage{hyperref}       % hyperlinks
\usepackage{url}            % simple URL typesetting
\usepackage{booktabs}       % professional-quality tables
\usepackage{amsfonts}       % blackboard math symbols
\usepackage{nicefrac}       % compact symbols for 1/2, etc.
\usepackage{microtype}      % microtypography
\usepackage{xcolor}         % colors
\usepackage{wrapfig}

\usepackage{graphicx}       % can i do it?
\title{Reinforcement Learning on Encrypted Data}

\author{%
  Alberto Jesu$^1$\thanks{Corresponding author. Email: \texttt{alberto.jesu@studio.unibo.it}},\enspace Victor-Alexandru Darvariu$^{2,3}$,\enspace Alessandro Staffolani$^1$,\\
  \textbf{Rebecca Montanari$^1$},\enspace \textbf{Mirco Musolesi$^{1,2,3}$} \\
  $^1$University of Bologna\quad$^2$University College London\quad $^3$The Alan Turing Institute\quad  \\
}

\begin{document}

\maketitle

\begin{abstract}
    % Reinforcement learning (RL) is a particular paradigm of machine learning that, recently, has proved to be a powerful approach, justifying its employment on a wide variety of use cases such as self-driving cars, games, and healthcare. Despite its success, however, in some scenarios the sensitive nature of the data required for the models to work arises privacy concerns that hinder the application of such methods. In this work, we explore the possibility of implementing a RL model in a privacy preserving environment by using encrypted states. We identify a state processing pipeline, and we train a Deep Q-Learning agent on encrypted versions of wide-spread benchmarking environments, assessing its ability to converge to a solution and the impact of the encryption step on its performance. The results highlight that the agent is still capable of learning in small state-spaces even in presence of non-deterministic encryption, but its generalizing capabilities on more complex environments suffers greatly.
    
The growing number of applications of Reinforcement Learning (RL) in real-world domains has led to the development of privacy-preserving techniques due to the inherently sensitive nature of data. Most existing works focus on differential privacy, in which information is revealed in the clear to an agent whose learned model should be robust against information leakage to malicious third parties. Motivated by use cases in which only encrypted data might be shared, such as information from sensitive sites, in this work we consider scenarios in which the inputs themselves are sensitive and cannot be revealed. We develop a simple extension to the MDP framework which provides for the encryption of states. We present a preliminary, experimental study of how a DQN agent trained on encrypted states performs in environments with discrete and continuous state spaces. Our results highlight that the agent is still capable of learning in small state spaces even in presence of non-deterministic encryption, but performance collapses in more complex environments.
    
\end{abstract}

\section{Introduction}
    % What could be a good structure here, paragraph-by-paragraph?
    % 1. RL and need for privacy / confidentiality in e.g. medical / search and rec applications. Focus on DP.
    % 2. Motivating example: IC systems: do not reveal state explicitly. Homomorphic encryption lets you achieve this. ML-confidential and cryptonets.
    % 3. Other related work: cryptographic operations / scrambling / data augmentation.
    % 4. Contribution: MDP extension, evaluation in different benchmarks / environments.

    Reinforcement Learning (RL) is a generic, reward-driven paradigm for learning complex behavioral policies through experience~\cite{bartosutton}. Certain real-world domains in which it has found success, such as healthcare~\cite{gottesman2019guidelines}, come with desiderata regarding the privacy of users due to the sensitivity of the data provided to the learning agent. Recent work adopts the differential privacy criterion~\cite{dwork2006calibrating}, injecting noise in order to protect quantities of interest from information leakage. These include work on policy evaluation~\cite{balle2016differentially}, Q-learning with function approximation~\cite{wang2019privacy}, learning a differentially private critic for transfer learning~\cite{actor-critic-DP}, and devising algorithms for the tabular case with PAC and regret guarantees~\cite{pac-and-regret}. Such work focuses on the use cases in which data \textit{can be revealed in the clear to the agent during the training phase}, but must be protected from leakage to malicious third parties (e.g., via probing~\cite{pan2019you}) once the policy is deployed.
    
    In certain cases, however, \textit{operating on encrypted training data} is desirable. One of the motivating examples of this work is that of control systems for which RL methods are attractive when dealing with nonlinear and stochastic dynamics~\cite{bucsoniu2018reinforcement}. In the case of industrial systems (e.g., manufacturing sites) and critical infrastructure (e.g., power plants) which are increasingly remotely controlled~\cite{patton2007generic}, sensitive information such as their layout and internal characteristics can be kept confidential through encryption. In recent years, Homomorphic Encryption (HE) schemes that allow for meaningful operations on ciphertexts~\cite{homomorphic:1, homomorphic:2, homomorphic:3, homomorphic:4, homomorphic:5} have led to the development of linear~\cite{MLConfidential} and deep neural network~\cite{DcryptoDL, cryptonets:2} classifiers in the supervised learning setting. However, to the best of our knowledge, there has not been work in RL that operates directly on ciphertexts, limiting applicability in such scenarios.
    
    While not directly comparable to cryptographic operations, there have been experimental studies that showed that sufficiently deep neural networks are able to easily fit image data in which pixels are shuffled or even completely replaced with noise~\cite{understandingdeep}. In the RL setting, works that use data augmentation~\cite{cobbe2019quantifying,laskin2020reinforcement}, random convolutional networks to output randomized feature maps~\cite{lee2020network}, or regularization of the Q-function in conjunction with randomly shifting the input image~\cite{kostrikov2020image} have all shown gains over standard algorithms. Furthermore, a concurrent work that studies shuffling inputs~\cite{tang2021sensory} shows that this approach brings improvements in policy robustness. %All in all, these works suggest that adding noise to RL input is not only viable approach, but can improve performance under certain circumstances.
    These results raise interesting questions around the possibility of applying RL on encrypted data. 
    
    % I would substitute -if noise- into the suggested words because encryption does not really produce noise (differential privacy techniques are more appropriate in that menaing)

    In this paper, we present an initial exploratory study of applying RL algorithms directly on ciphertexts. We contribute a simple extension to the MDP framework and DQN algorithm that provides the learning agent with encrypted state information. We perform an empirical study of various encryption schemes in two distinct environments, with states and actions that are discrete and continuous respectively. We explore the agents' abilities to learn in these settings and we analyze several design and implementation aspects, such as the choice of encryption scheme, as well as key length and padding in the case of block cipher modes. We identify scenarios in which a solution based on encryption achieves comparable results to using plaintext training data. Finally, we discuss the inherent challenges and opportunities of using these techniques in more complex environments.

\section{Approach}
    \label{approach}
    \begin{figure}[t]
    \centering
    \includegraphics[width=\textwidth]{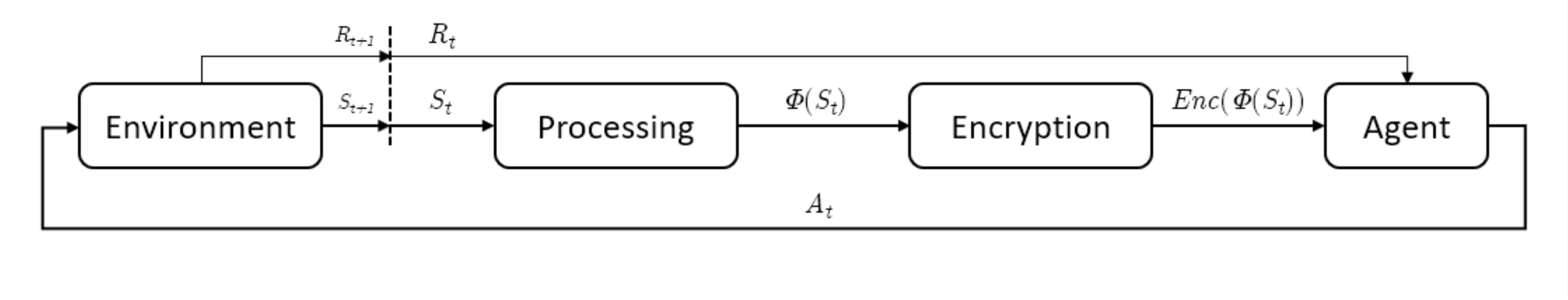}
    \caption{Illustration of the extended MDP formulation, in which states are processed and encrypted prior to being fed to the agent.}
    \label{fig:general-workflow}
    \end{figure}
    %
    % Being able to apply RL to encrypted states would mean creating an agent that learns a policy without supplying it with complete information about its surroundings. This would allow us to decouple the environment in which the agent moves and acts from the algorithms that control such agent, leading the way to numerous use cases similar to the ones explained in the previous papers.
    %
    Our extended MDP formulation, illustrated in Figure~\ref{fig:general-workflow}, consists of four main components: the \emph{environment} with which the agent interacts, a \emph{state processing step}, which applies an environment-dependent processing function (or compositions of functions) $\phi(S_t)$ to the state $S_t$ at time $t$; an \emph{encryption step} with a given encryption primitive $Enc(\cdot)$; and an \emph{agent} that receives the transformed and encrypted states $Enc(\phi(S_t))$ as well as the plaintext reward $R_t$. For example, in our experiments, in the processing steps, we applied functions for resizing state images, adding padding for block ciphers, and performing other pre-processing operations.
    
    \textbf{Learning algorithm.} We opt for extending the DQN~\cite{dqn:2} algorithm over policy gradient methods due to its sample efficiency. Specifically, we include an additional layer of state manipulation that preprocesses and encrypts the state. In this way, the original information provided in the state is modified so that it is not accessible to the agent and any third party that may be able to intercept communications via this channel, which is considered insecure by default. The states are stored in the replay memory in the encrypted version and are randomly sampled to perform the updates of the neural network using the Q-targets.
    
    \textbf{Encryption primitives.} We consider several encryption primitives $Enc(\cdot)$. In terms of block ciphers, we opt for  AES~\cite{AES} in ECB mode and CBC mode (noting that the former is cryptographically insecure exhibiting deterministic properties). For these ciphers, plaintexts need to be padded such that they become a multiple of the block size. We also consider the CKKS~\cite{ckks} homomorphic encryption scheme, widely used in machine learning applications. As baselines, we also include a no-op primitive that outputs the plaintext as well as a deterministic shuffling of the pixels of the state for environments with image inputs, which applies a consistent permutation to the pixels of an image.

\section{Experimental Setup}
\label{exps}
We assess the ability of our agent to learn on encrypted data by conducting two case studies on OpenAI Gym~\cite{Gym} environments: MiniGrid \cite{Gym_minigrid}, a discrete navigation task, and LunarLander, a continuous control task. We implement the extended MDP framework and DQN algorithm outlined in the previous section, and train an RL agent multiple times over states encrypted with different encryption schemes and compare the results at the end of the training run. The training is repeated across 10 different seeds for statistical validity. The homomorphic encryption functions are implemented in Pyfhel~\cite{Pyfhel}, a Python wrapper around Microsoft SEAL~\cite{sealcrypto}, while for block ciphers we use the \texttt{pyca/cryptography}~\cite{pyca} module.

\begin{figure}[t]
    \centering
    \includegraphics[width=\textwidth]{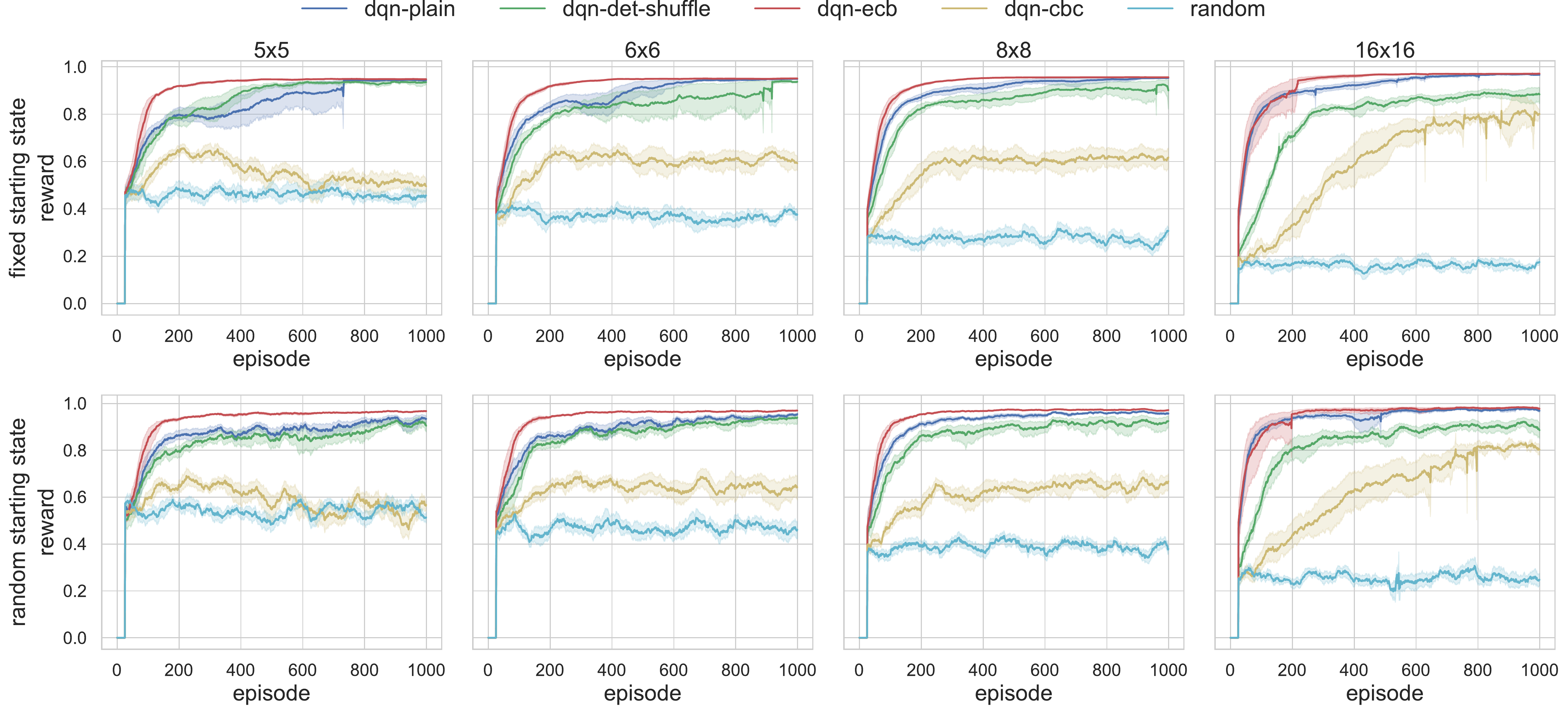}
    \caption{Training performance in MiniGrid.}
    \label{fig:train-minigrid}
\end{figure}

\textbf{MiniGrid.} In the first case study, the environments chosen for the experiments are the empty room scenarios from the MiniGrid collection, both in the fixed starting state and random starting state variants. We consider the $5\times5$, $6\times6$, $8\times8$ and $16\times16$ grid sizes. The observations are $\langle H \times W \times 3 \rangle$ RGB images of the room, which are then processed by the preprocessing function $\phi(\cdot)$ via a resizing of the image to 1 pixel per tile, a greyscale conversion and the cropping of outer walls. The direction of the agent, which is lost during resizing, is then re-established by remapping the pixel corresponding to the agent to one of four predetermined pixel intensities. The Q-network is parametrized by a Convolutional Neural Network (CNN).
\begin{wrapfigure}{r}{0.45\textwidth}
    \vspace{-\baselineskip}
    \begin{center}
        \includegraphics[width=0.45\textwidth]{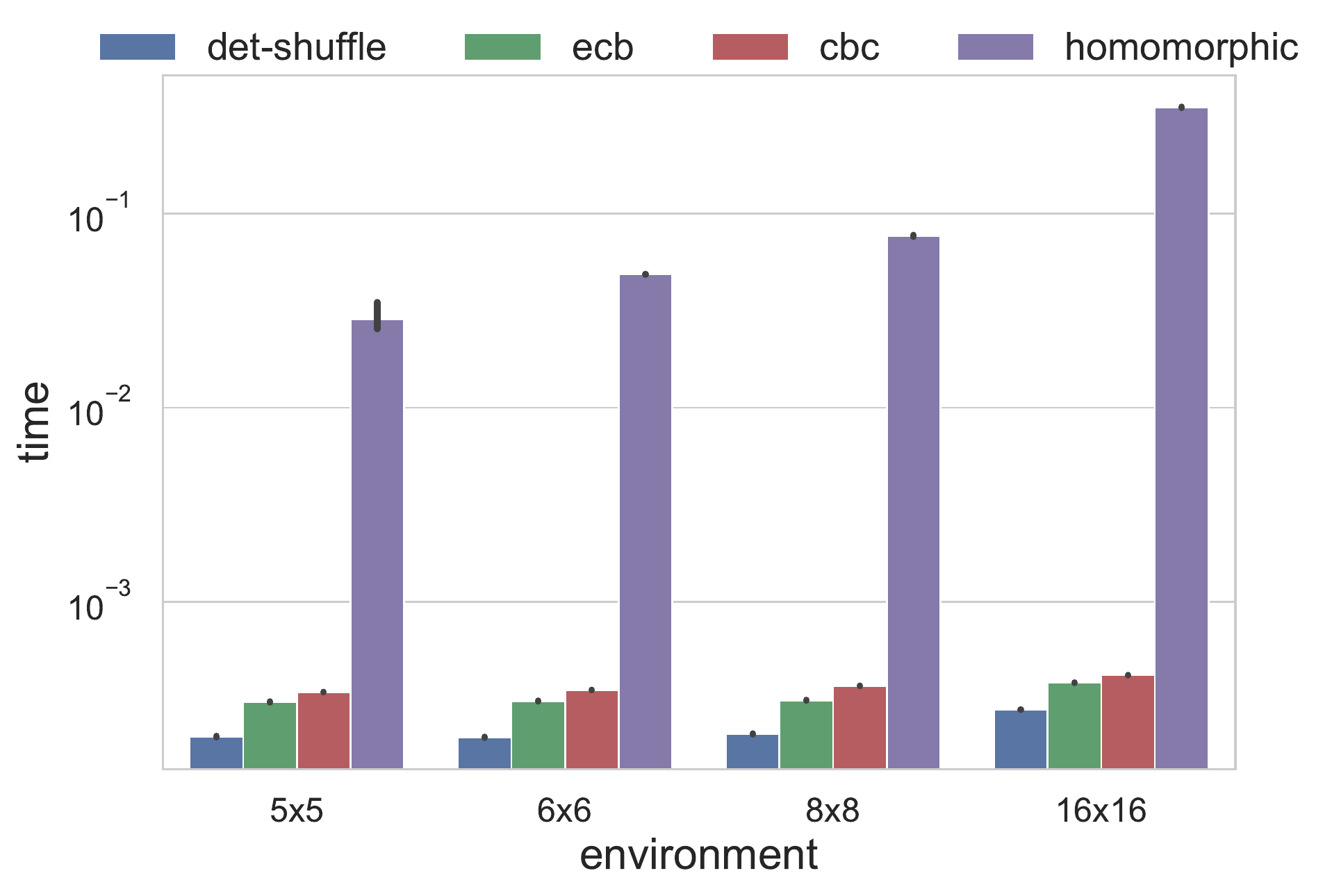}
        \vspace{-\baselineskip} 
        \caption{Encryption time (in seconds).}
        \label{fig:enc-times}
    \end{center}
    \vspace{-\baselineskip}
\end{wrapfigure}
For block ciphers, the images also need to be padded before the encryption step. We consider two padding schemes: \textit{custom} and the \textit{PKCS\#7} standard. The first approach pads the images by adding rows and columns, enough to reach a multiple of the block size, of pixels with a specific intensity. This approach allows to maintain the spatial distribution of the pixels when the ciphertexts are passed to the CNN and the bytes are reassembled in a matrix-like configuration for the forward pass in the neural network. PKCS\#7 adds $k - (l |k|)$ bytes at the trailing end of the plaintext (where $l$ is the length of the input message, and $k$ is the block size), resulting in an additional layer of pixel scrambling when a bidimensional input is reassembled from the ciphertext. 

\textbf{LunarLander.} The second case study was conducted on LunarLander, a continuous control task, in which observations are 8-dimensional arrays indicating positional information. Since directly encrypting these values would result in far too many possible ciphertexts, the preprocessing function $\phi(S_t)$ consists in the discretization of the continuous parts of the state in 256 bins -- which serves as a compromise between granularity and feasibility of working with encrypted states.
The encryption functions over which the agent is trained are the same of the previous case study except for the deterministic shuffling, which would only rearrange the variables of the array. The padding technique considered is PKCS\#7, and the Q-function is parametrized by an MLP.

%In the first case study, the environments chosen for the experiments are the empty room scenarios from the MiniGrid %collection, in the fixed starting state and random starting state variants. The observations are $\langle H \times W %\times 3 \rangle$ RGB images of the room. The preprocessing function $\phi(S_t)$ consists in the resizing of the %image to 1 pixel per square, a greyscale conversion and the cropping of the outer walls. The direction of the agent, %which is lost in the process, is then re-established by remapping the pixel corresponding to the agent with one of %four predefined pixel intensities. Lastly, the agent is trained four times over different encryption functions: a %plaintext case, which acts as a baseline; a deterministic shuffling of the input pixels; AES in ECB mode; and AES in %CBC mode. The performanc Since AES is a block cipher, the images also undergo padding before encryption. The %performances of the agent are then compared, using the plaintext case as a baseline; furthermore, the impact of the %key size and padding technique are also investigated. Figure \ref{fig:train-minigrid} shows the results of the %experiments,

%
\begin{wraptable}{R}{0.47\textwidth}
\vspace{-1.5\baselineskip}
\caption{Reward at convergence with different key sizes (top) and padding schemes (bottom).}\label{tab:keysize}
%\vspace{\baselineskip}
\resizebox{0.47\textwidth}{!}{
\begin{tabular}{lcccc}\\
Encryption & 5x5 & 6x6 & 8x8 & 16x16 \\\midrule
ECB(32) & 0.948\tiny{$\pm0.001$} & 0.950\tiny{$\pm0.000$} & 0.955\tiny{$\pm0.000$} & 0.898\tiny{$\pm0.107$} \\
ECB(24) & 0.948\tiny{$\pm0.001$} & 0.950\tiny{$\pm0.000$} & 0.956\tiny{$\pm0.000$} & 0.918\tiny{$\pm0.081$} \\
ECB(16) & 0.949\tiny{$\pm0.001$} & 0.950\tiny{$\pm0.000$} & 0.956\tiny{$\pm0.000$} & 0.962\tiny{$\pm0.013$} \\
CBC(32) & 0.481\tiny{$\pm0.029$} & 0.544\tiny{$\pm0.033$} & 0.614\tiny{$\pm0.028$} & 0.729\tiny{$\pm0.066$} \\
CBC(24) & 0.490\tiny{$\pm0.019$} & 0.554\tiny{$\pm0.018$} & 0.661\tiny{$\pm0.027$} & 0.748\tiny{$\pm0.055$} \\
CBC(16) & 0.489\tiny{$\pm0.019$} & 0.549\tiny{$\pm0.021$} & 0.664\tiny{$\pm0.030$} & 0.698\tiny{$\pm0.062$} \\\midrule
Plain + Custom & 0.905\tiny{$\pm0.092$} & 0.949\tiny{$\pm0.001$} & 0.955\tiny{$\pm0.001$} & 0.934\tiny{$\pm0.028$}\\
Plain + PKCS & 0.913\tiny{$\pm0.0714$} & 0.921\tiny{$\pm0.057$} & 0.914\tiny{$\pm0.071$} & 0.927\tiny{$\pm0.039$} \\
Shuffle + Custom & 0.943\tiny{$\pm0.005$} & 0.887\tiny{$\pm0.082$} & 0.912\tiny{$\pm0.049$} & 0.802\tiny{$\pm0.159$} \\
Shuffle + PKCS & 0.943\tiny{$\pm0.004$} & 0.946\tiny{$\pm0.004$} & 0.952\tiny{$\pm0.001$} & 0.966\tiny{$\pm0.008$} \\
ECB + Custom & 0.948\tiny{$\pm0.001$} & 0.950\tiny{$\pm0.000$} & 0.955\tiny{$\pm0.000$} & 0.898\tiny{$\pm0.107$} \\
ECB + PKCS & 0.949\tiny{$\pm0.001$} & 0.950\tiny{$\pm0.001$} & 0.955\tiny{$\pm0.000$} & 0.971\tiny{$\pm0.000$} \\
CBC + Custom & 0.481\tiny{$\pm0.029$} & 0.544\tiny{$\pm0.033$} & 0.614\tiny{$\pm0.028$} & 0.730\tiny{$\pm0.066$} \\
CBC + PKCS & 0.481\tiny{$\pm0.026$} & 0.548\tiny{$\pm0.030$} & 0.639\tiny{$\pm0.035$} & 0.701\tiny{$\pm0.065$} \\\bottomrule
\end{tabular}
}
\vspace{-\baselineskip}
\end{wraptable}

\section{Results}
\label{res}

\textbf{Overhead of encryption step.} Since the protocol behind training an RL algorithm involves constant interaction with the environment, we  conduct a preliminary experiment in order to quantify the overhead introduced by the encryption part of the state processing pipeline. We measure the elapsed wall clock time taken by various schemes to encrypt a single state in the MiniGrid environments, repeating this measurement $1000$ times. Results are shown in Figure~\ref{fig:enc-times}. We find that block ciphers introduce a negligible overhead compared to a simple shuffling. In contrast, performing a HE step is two to three orders of magnitude slower, rising with the state size. We conclude that the time complexity of current HE techniques, coupled with the sample complexity of current RL algorithms, make this approach unfeasible. Therefore, we do not consider HE techniques in subsequent experiments.

%One of the first step performed to evaluate the agent over encrypted data is to measure the overhead caused by the encryption step during the state processing pipeline. Since reinforcement learning is a very interactive problem, a big encryption overhead would reduce drastically the feasibility of this approach. These results are compared to a sample of a homomorphic encryption approach, typical of the papers discussed. As the Table suggests, the encryption step for the considered cryptographic functions does not induce a significant overhead. On the other hand, we can see that homomorphic encryption's time complexity, with respect to AES, is hundreds, or thousands for the biggest state, times slower. Thus, homomorphic encryption, as of now, does not represent a feasible encryption method for reinforcement learning problems.

\textbf{MiniGrid performance.} Figure \ref{fig:train-minigrid} shows the comparison of the results across different encryption schemes and environment sizes, using a key length of $32$ bytes and the custom padding scheme. Our results highlight that the agent is able to converge to an optimal solution in the plaintext case, the deterministic shuffle case, and the AES in ECB mode case in both the fixed starting state and random starting state scenarios, due to the deterministic nature of the transformations. Interestingly, applying AES with ECB leads to a gain in performance in all settings tested. However, for the AES in CBC mode case, despite not being able to reach an optimal solution, the agent is still capable of outperforming the random baseline, which is more apparent as environment size increases.
\begin{wrapfigure}{r}{0.40\textwidth}
    \vspace{-\baselineskip}
    \begin{center}
        \includegraphics[width=0.40\textwidth]{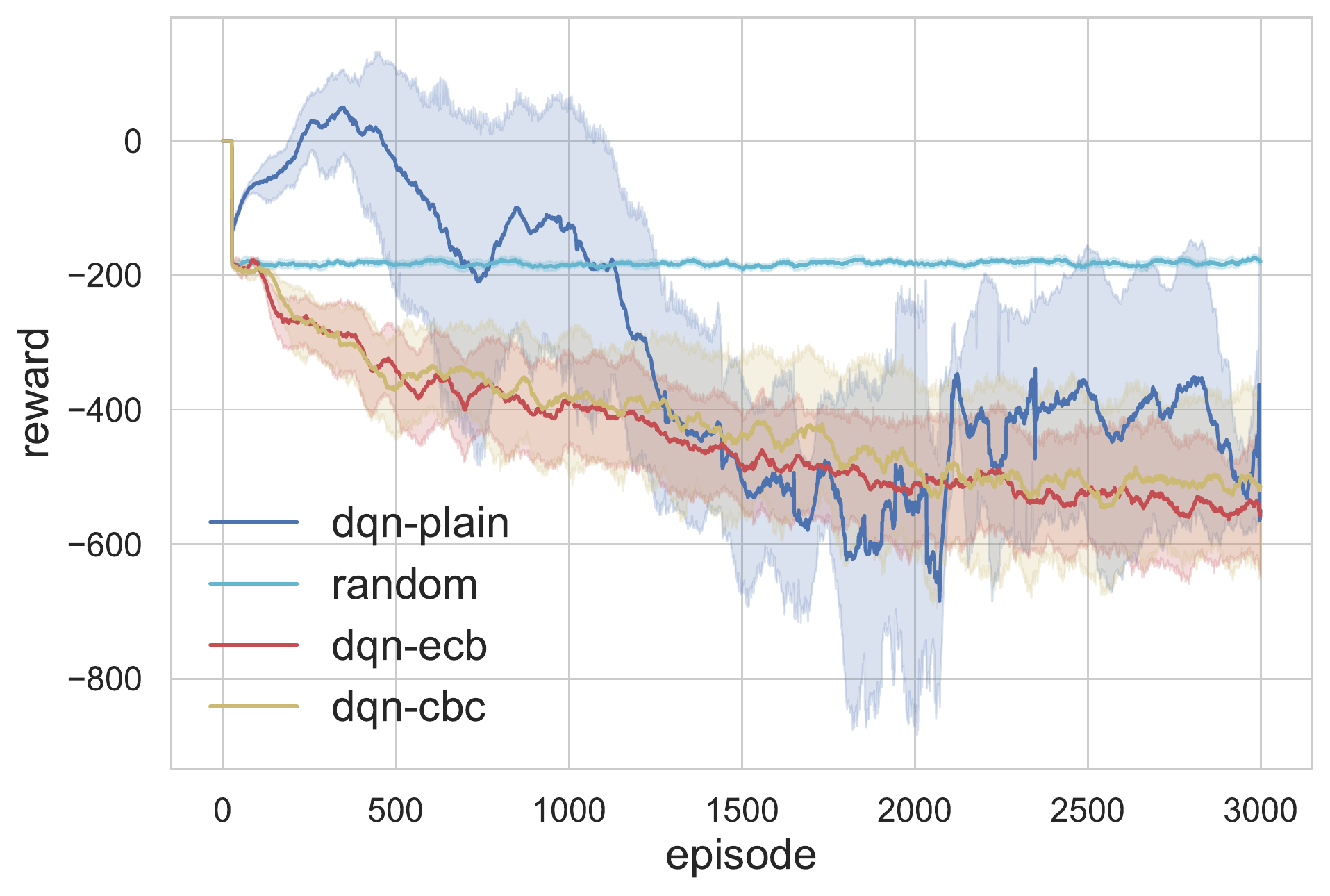}
        \vspace{-\baselineskip} 
        \caption{Training performance in LunarLander.}
        \label{fig:train-lunarlander}
    \end{center}
    \vspace{-\baselineskip}
\end{wrapfigure}

\textbf{Impact of padding and key length.} Table \ref{tab:keysize} shows the average reward at convergence in the fixed starting state environments using variable key lengths for the block cipher: $32$, $24$, and $16$ bytes respectively. As reported, the key size does not induce a significant difference, most likely due to the small state space of the environment. While a bigger key increments the space of possible ciphertexts, the possible amount of states is bounded by the small dimensions of the environments and of the images. The two padding techniques taken into consideration do not lead to a statistically significant difference in results. %Similarly, no statistically significant difference was observed by using two different padding techniques. %, despite the custom technique being engineered specifically for this application.

\textbf{LunarLander performance.} Figure \ref{fig:train-lunarlander} shows the training performance of the agent with this setup. In the plaintext case, we observe an initial learning period, followed by a drastic divergence. No learning appears to take place and the performance is progressively worse. %The encrypted versions do not show any trace of learning, with their performance becoming progressively worse.
We hypothesize that, while for the discrete action environments ``memorizing" the optimal path is easy, for continuous action space environments even small variations in outputs may quickly lead the agent to unexplored parts of the state space. It is worth noting that, even with the binning used, the state space has a size of $2^{64}$.

\section{Conclusions}
In this work, we have presented an exploratory study to assess the feasibility of applying RL on encrypted data. We have introduced a simple extension to the MDP framework and the DQN algorithm, and presented two case studies which assessed the impact of various encryption schemes on the agents' learnability. Firstly, our findings suggest that applying HE in conjunction with RL is unrealistic on commodity hardware, while block ciphers introduce a small overhead. Secondly, for small state spaces and discrete actions, deterministic transformations perform on par with a plaintext agent and, quite surprisingly, a non-deterministic transformation (AES+CBC) outperforms randomly chosen actions significantly. Finally, the generalization capabilities suffer as the state space becomes large and control signals continuous, a regime in which none of the approaches succeeded. %We hope this work can represent an initial step towards the design of efficient algorithms for RL on encrypted data.

%In this work, we have assessed the feasibility of the application of RL over encrypted data. We have extended the classic MDP framework and DQN algorithm so as to include encrypted states, and defined a set of preprocessing functions and padding techniques to prepare the states for the encryption. Lastly, we have presented two case studies in which we have applied our algorithm to different environments and studied the impact of the transformations on the ability to learn of the agent. From the results, it emerges that the agent is able to converge to an optimal solution in presence of deterministic transformations and small state spaces, while its generalisation capabilities suffer as the state space becomes larger. 
\bibliography{./biblio-filtered.bib}
% is it possible to extend a little bit the argumentation of the difficulty of generalisation?  
%%%%%%%%%%%%%%%%%%%%%%%%%%%%%%%%%%%%%%%%%%%%%%%%%%%%%%%%%%%%

\end{document}